\documentclass[letterpaper, 10 pt, conference]{ieeeconf}  % Comment this line out if you need a4paper

\IEEEoverridecommandlockouts                              % This command is only needed if 
\usepackage{graphics} % for pdf, bitmapped graphics files
\usepackage{epsfig} % for postscript graphics files
\usepackage{mathptmx} % assumes new font selection scheme installed
\usepackage{times} % assumes new font selection scheme installed
\usepackage{amsmath} % assumes amsmath package installed
\usepackage{amssymb}  % assumes amsmath package installed
\usepackage{xcolor}
\usepackage{caption}
\usepackage{subcaption} %For sub-figures
\usepackage{gensymb}
\usepackage{cite}

\renewenvironment{align*}{\align}{\endalign}

\title{\LARGE \bf
Unsupervised Odometry and Depth Learning for Endoscopic Capsule Robots
}

\author{Mehmet Turan$^{1}$, Evin Pinar Ornek$^{2}$, Nail Ibrahimli $^{2}$, Can Giracoglu$^{2}$, Yasin Almalioglu$^{3}$, \\
Mehmet Fatih Yanik $^{4}$, and Metin Sitti$^{5}$% <-this % stops a space
\thanks{$^{1}$Mehmet Turan is with Physical Intelligence Department, Max Planck Institute for Intelligent Systems, Germany 
        {\tt\small turan@is.mpg.de}}%
\thanks{$^{2}$Evin Pinar Ornek, Nail Ibrahimli, Can Giracoglu is with the Informatics Faculty, Technical University of Muenich, Germany
        {\tt\small evin.oernek, nail.ibrahimli, can.giracoglu@tum.de}}%     
\thanks{$^{3}$Yasin Almalioglu is with Computer Science Department, University of Oxford, Oxford, UK
        {\tt\small yasin.almalioglu@cs.ox.ac.uk}}%
        \thanks{$^{4}$M. Fatih Yanik is with the Department of Information Technology and Electrical Engineering, Zurich, Switzerland
        {\tt\small yanik@ethz.ch}}%
\thanks{$^{5}$Metin Sitti is with Physical Intelligence Department, Max Planck Institute for Intelligent Systems, Germany
        {\tt\small sitti@is.mpg.de}}%
}

\begin{document}

\maketitle
\thispagestyle{empty}
\pagestyle{empty}

%%%%%%%%%%%%%%%%%%%%%%%%%%%%%%%%%%%%%%%%%%%%%%%%%%%%%%%%%%%%%%%%%%%%%%%%%%%%%%%%
\begin{abstract}
In the last decade, many medical companies and research groups have tried to convert passive capsule endoscopes as an emerging and minimally invasive diagnostic technology into actively steerable endoscopic capsule robots which will provide more intuitive disease detection, targeted drug delivery and biopsy-like operations in the gastrointestinal(GI) tract. In this study, we introduce a fully unsupervised, real-time odometry and depth learner for monocular endoscopic capsule robots. We establish the supervision by warping view sequences and assigning the re-projection minimization to the loss function, which we adopt in multi-view pose estimation and single-view depth estimation network. Detailed quantitative and qualitative analyses of the proposed framework performed on non-rigidly deformable ex-vivo porcine stomach datasets proves the effectiveness of the method in terms of motion estimation and depth recovery.
\end{abstract}

%%%%%%%%%%%%%%%%%%%%%%%%%%%%%%%%%%%%%%%%%%%%%%%%%%%%%%%%%%%%%%%%%%%%%%%%%%%%%%%%

\section{Introduction} \label{sec:intro}

Advancements in various fields of science and technology in the last decade has opened new pathways for non-invasive examination of patient's body and detailed investigation about diseases. Hospitals are using innovative ways to provide accurate data from inside of the human body.
As an emerging example, various diseases such as colorectal cancer and inflamatory bowel disease are diagnosed by the  usage of swallowable capsule endoscopes, which are non-invasive, painless, suitable to be used for long duration screening purposes which can access difficult body parts (e.g.,small intestines) better than standard endoscopy. Such benefits make swallowable, non-tethered capsule endoscopes an exciting alternative over standard endoscopy \cite{sitti2015biomedical, turan2017endovo}. 

\begin{figure}[t]

	\begin{subfigure}[t]{0.4\textwidth} 
	    \includegraphics[width=1\linewidth]{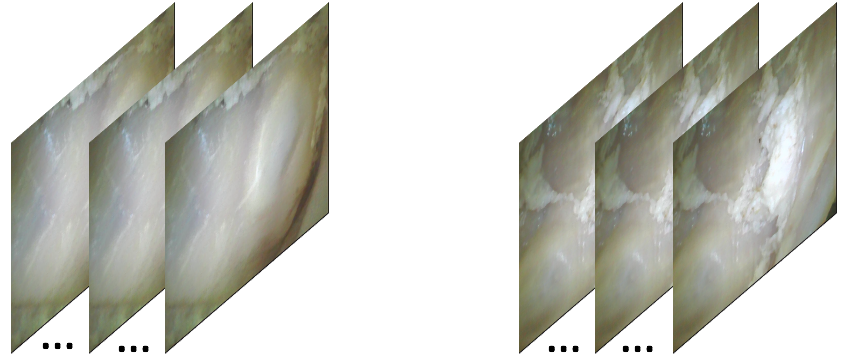}
	    \caption{Training: Unlabeled image sequences}
	    \label{fig:train}
	\end{subfigure}
	
	\begin{subfigure}[t]{0.4\textwidth} 
	    \includegraphics[width=1\linewidth]{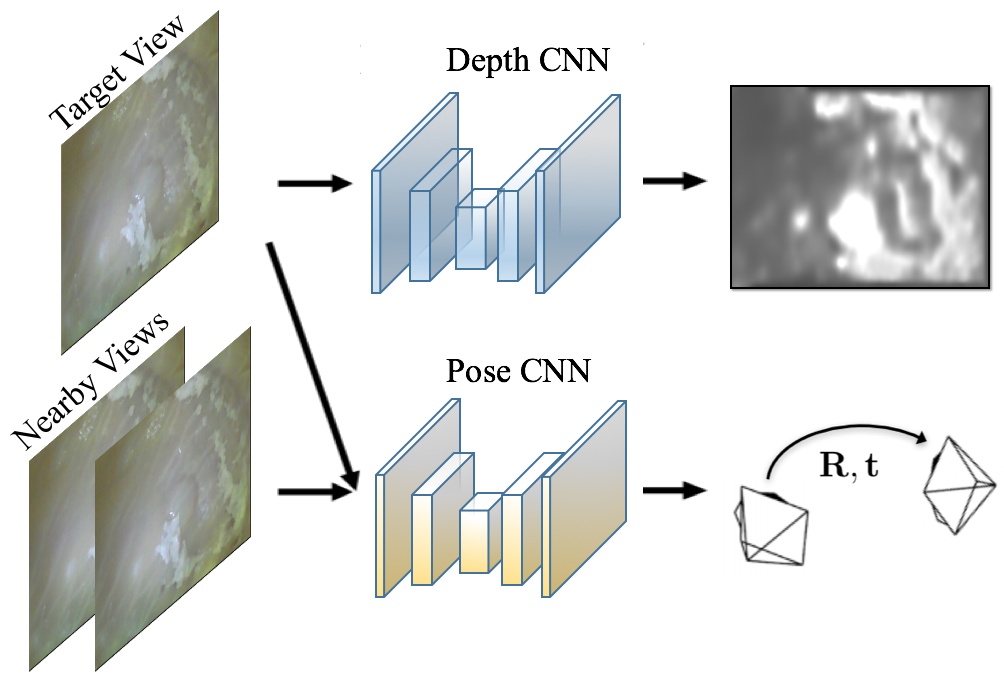}
	    \caption{Testing: Pose and depth prediction}
	    \label{fig:test}
	\end{subfigure}

    \setlength{\belowcaptionskip}{-14pt}
	\caption{Unsupervised training approach consists of two separate neural networks, one for depth prediction and another one for multi-view pose estimation. It requires unlabeled image sequences from different temporal points to establish a supervision basis. Models produce pose estimation between two views from different perspectives parameterized as 6-DoF motion, and depth prediction as a disparity map for a given view. } 
	\label{fig:networks} 
	
\end{figure}

Current capsule endoscope technology employed in GI tract monitoring and disease detection consists of passive devices which are locomated by random peristaltic motions.
The doctor would have an easier access to fine-scale body parts and could make more intuitive and correct diagnosis in case of a precise and reliable control over the position of the capsule. Many research groups attempted to build remotely controllable active endoscopic capsule robot systems with additional functionalities such as local drug delivery, biopsy and other medical functions \cite{goenka2014capsule, turanendovmfuse2017, TURAN20181861, turanendosensor2017, Turan2017, Turan2018, turan2017endovo, nakamura2008capsule, munoz2014review, carpi2011magnetically, keller2012method, mahoney2013managing, yim2014biopsy, petruska2013omnidirectional, DBLP:journals/corr/TuranAKS17, DBLP:journals/corr/TuranPJAKS17, DBLP:journals/corr/TuranAJAKS17, DBLP:journals/corr/TuranAAKS17  }, which are, on the other hand, heavily dependent on a real-time and precise pose estimation capability.

\begin{figure}[t]
	\includegraphics [trim=0 0 0 0, clip, angle=0, width=\columnwidth, keepaspectratio]{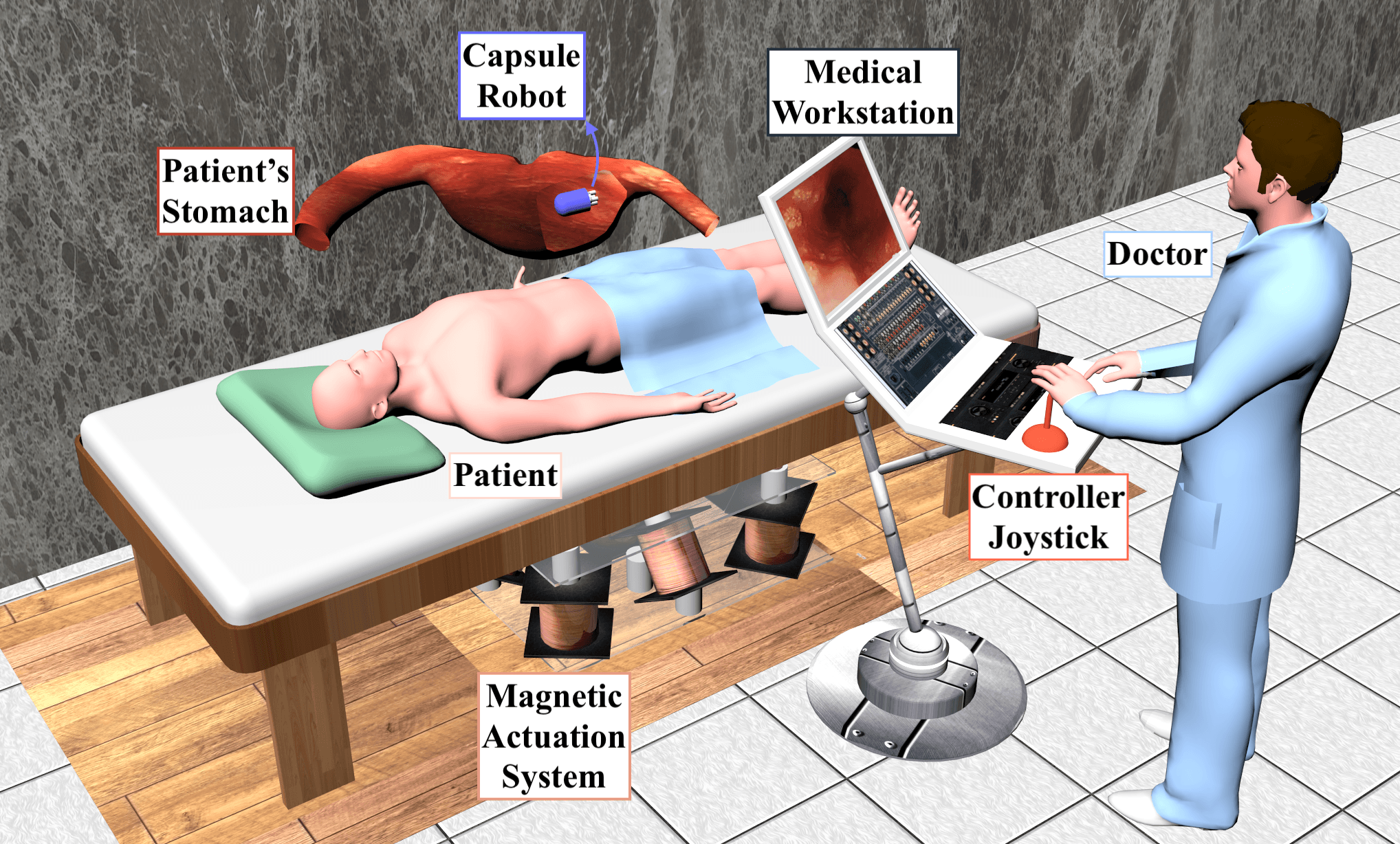}
	\setlength{\belowcaptionskip}{-10pt}
	\caption{Demonstration of the active endoscopic capsule robot operation using MASCE (Magnetically actuated soft capsule endoscope) designed for disease detection, drug delivery and biopsy-like operations in the upper GI-tract. MASCE is composed of a RGB camera, a permanent magnet, an empty space for drug chamber and a biopsy tool. Electromagnetic coils based actuation unit below the patient table exerts forces and torques to execute the desired motion. Doctor operates the screening, drug delivery and biopsy processes in real-time using the live video stream onto the medical workstation and the controller joystick to maneuver the endoscopic capsule to the desired position/orientation and to execute desired therapeutic actions such as drug release and biopsy.}
    \label{fig:goal}
\end{figure}

In this work, we propose a novel real-time localization and depth estimation approach for endoscopic capsule robots which mimic the remarkable ego-motion estimation and scene reconstruction capabilities of human beings by training an unsupervised deep neural network. The proposed network consists of two simultaneously trained sub networks, the first one assigned for depth estimation via encoder-decoder strategy, the second assigned to regress the camera pose in 6-DoF. The model observes sequences of monocular images and aims to interpret them to estimate executed camera motion in 6-DoF and the depth map of the observed scene as shown in Fig. \ref{fig:networks}. Our framework estimates the camera motion and depth information in an end-to-end and unsupervised fashion directly from input pixels. Training is performed using only unlabeled monocular frames in a similar way to prior works such as \cite{szeliski1999prediction, zhou2017unsupervised, flynn2016deepstereo}. 

We formulate the entire pose estimation and map reconstruction pipeline for endoscopic capsule robots as a consistent and systematic learning concept which can improve its performance every day by collecting streamed data belonging to numerous patients undertaken to endoscopic capsule robot and standard endoscopy investigations in hospitals over the world. This way, we want to mimic and transfer a continuous learning functionality from medical doctors into medical robots domain, where experience and adaptation to unexpected novel situations can be much more critical to real-world scenarios. 

\begin{figure}[t]
	\centering
	\includegraphics [trim=0 0 0 0, clip, angle=0, width=\columnwidth, keepaspectratio]{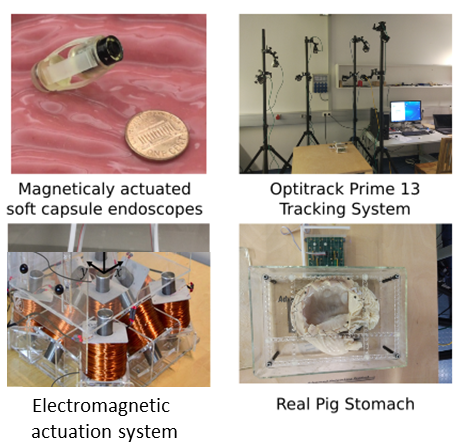}
	\setlength{\belowcaptionskip}{-12pt}
	\caption{Illustration of the experimental setup. MASCE is a magnetically actuated robotic capsule endoscope prototype which has a ringmagnet on the body. An electromagnetic coil array consisting of nine coils is used for the actuation of the MASCE. The ringmagnet exerts magnetic force and torque on the capsule in response to the external magnetic field provide by the electromagnetic coil array. Magnetic torque and forces are also used to release drug, as well. OptiTrack system consisting of eight infrared cameras is employed for the ground truth pose estimation. An opened and oiled porcine stomach simulator is used to represent human stomach.}
\label{fig:opti} 
\end{figure}

To summarize, main contributions of our paper are as follows: 
\begin{itemize}
	\item To best of our knowledge, this is the first unsupervised odometry and depth estimation approach for both the endoscopic capsule robots and hand-held standard endoscopes.
	\item Since the network learns in a fully unsupervised manner, no ground truth pose and/or depth values are required to train the neural network.
	\item Neither prior knowledge nor parameter tuning is needed to recover the trajectory and depth, contrary to traditional visual odometry(VO) and deep learning(DL) based supervised odometry approaches.
    \item We simultaneously train a reliability mask which identifies pixels distorted by camera occlusions, non-rigid organ deformations and/or non-Lambertian surface. Such a mask is very crucial for vision based methods applied on endoscopic type of images since occlusions, non-rigid deformations and specularities violating Lambertian surface properties commonly occur in endoscopic types of images.
\end{itemize}

Evaluations we made on non-rigidly deformable porcine stomach videos prove the success of our depth estimation and localization approach. As the outline of this paper, the previous work in endoscopic capsule odometry is discussed in Section \ref{sec:background}. Section \ref{sec:method} introduces the proposed method with its mathematical background in detail and the unsupervised DL architecture. Section \ref{sec:results} shows our experimental quantitative and qualitative results achieved for 6-DoF localization and depth recovery. Finally, Section \ref{sec:conclusion} mentions some bottlenecks and gives future directions for our project. Our code will be made available at https://github.com/mpi/deep-unsupervised-endovo.

%%%%%%%%%%%%%%%%%%%%%%%%%%%%%%%%%%%%%%%%%%%%%%%%%%%%%%%%%%%%%%%%%%%%%%%%%%%%%%%%

\section{Background} \label{sec:background}

\begin{figure}[t]
	\centering
	\includegraphics [trim=0 0 0 0, clip, angle=0,width=\columnwidth, keepaspectratio]{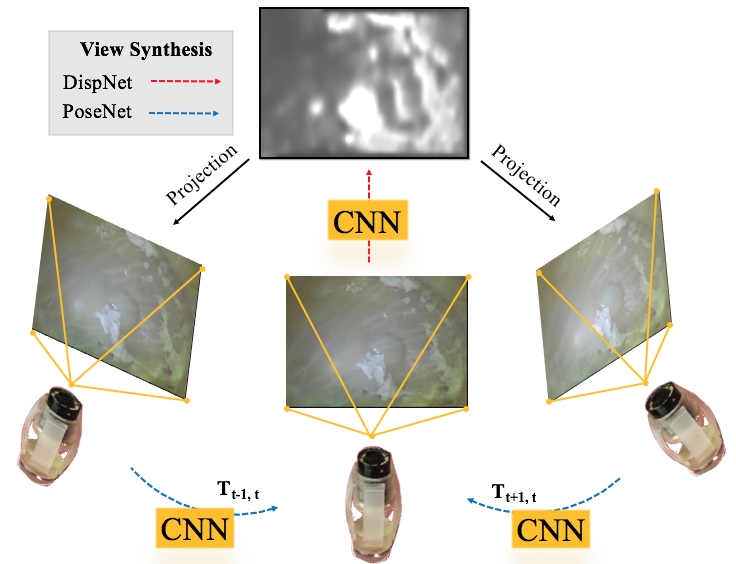}
	\caption{Training input consists of sequential images from different perspectives, which are noted by $<I_{t-1}, I_{t}, I_{t+1}>$. After view synthesis creates the supervision baseline, PoseNet is trained to estimate relative motion change between $<I_{t-1}, I_{t}>$ and $<I_{t}, I_{t+1}>$, whereas DispNet learns to predict depth for the target image $<I_{t}>$. 
	} 
	\label{fig:seq} 
\end{figure}

In the last decade, several localization methods \cite{than2012review, fluckiger2007ultrasound, rubin2006sonographic, kim2008noninvasive, yim20133} were proposed to calculate the 3D position and orientation of the endoscopic capsule robot such as fluoroscopy \cite{than2012review}, ultrasonic imaging \cite{fluckiger2007ultrasound, rubin2006sonographic, kim2008noninvasive, yim20133}, positron emission tomography (PET) \cite{than2012review, yim20133}, magnetic resonance imaging (MRI) \cite{than2012review}, radio transmitter based techniques and magnetic field based techniques. The common drawback of these localization methods is that they require extra sensors and hardware design. Such extra sensors have their own drawbacks and limitations if it comes to their application in small scale medical devices such as space limitations, cost aspects, design incompatibilities, biocompatibility issues and the interference of the sensors with the activation system of the device. 

As a solution of these issues, a trend of VO methods have attracted the attention for endoscopic capsule localization. A classic VO pipeline typically consists of many hand-engineered parts such as camera calibration, feature detection, feature matching, outliers rejection (e.g. RANSAC), motion estimation, scale estimation and global optimization (bundle adjustment). Although some state-of-the-art algorithms based on this traditional pipeline have been developed and proposed for endoscopic VO task in the past decades, their main deficiencies such as tracking failures in low textured areas, sensor occlusion issues, lack of handling non-rigid organ deformation still remain. In last couple of years, DL techniques have been dominating many computer vision related tasks with numerous promising result, e.g. object detection, object recognition, classification problems etc. Contrary to these high-level computer vision tasks, VO is mainly working on motion dynamics and relations across sequence of images, which can be defined as a sequential learning problem. 

Our proposed method solves several issues faced by typical VO pipelines, e.g the need to establish a frame-to-frame feature correspondence, vignetting artefacts, motion blur, specularity or low signal-to-noise ratio (SNR). We think that DL based endoscopic VO approach is more suitable for such challenge areas since the operation environment(GI tract) has similar organ tissue patterns among different patients which can be learned by a sophisticated machine learning approach easily. Even the dynamics of common artefacts such as non-rigidness, sensor occlusions, vignetting, motion blur and specularity across frame sequences could be learned and used for a better pose estimation, whereas our unsupervised odometry learning method additionally solves the common problem of missing labels on medical datasets from inner body operations \cite{turanendovmfuse2017, turanendosensor2017}.

%%%%%%%%%%%%%%%%%%%%%%%%%%%%%%%%%%%%%%%%%%%%%%%%%%%%%%%%%%%%%%%%%%%%%%%%%%%%%%%%

\section{METHOD} \label{sec:method}

\begin{figure*}
\includegraphics[width=\textwidth]{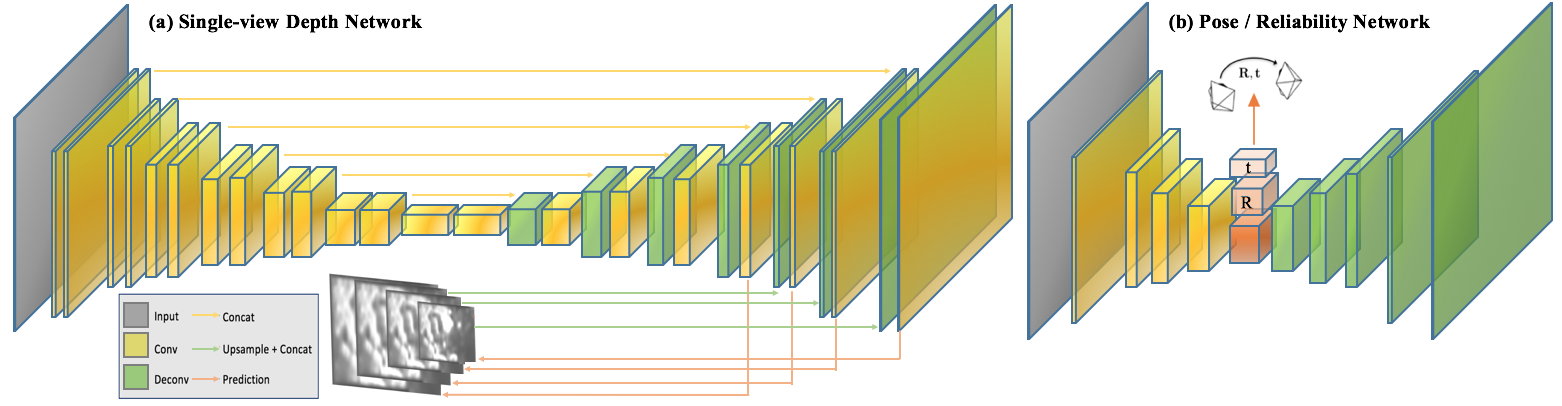}
\caption{The proposed neural network architecture for pose/reliability/depth map estimation. The width and height of illustrated blocks reflect the spatial dimensions of layers and output channels which are based on an encoder-decoder design. (a) Single-view depth prediction model is adopted by DispNet\cite{mayerIHFCDB15}. ReLu activations follow the middle convolution layers. Kernel size for first four layers are 7, 7, 5, 5 respectively, and rest of the layers have kernel size 3. (b) Pose/reliability estimation network is motivated by SFM-Learner\cite{zhou2017unsupervised} model and it has decoder-encoder design, as well. The encoder part has five feature extraction layers which are shared for both pose and reliability mask estimation. The pose results are gathered after the encoder network, which has $6*(N-1)$ output channels for 6-DoF motion parameters. The encoder part is followed by a decoder, which has 5 deconvolutional layers, consisting ReLU activations in between. }
\label{fig:deeparchitecture}
\end{figure*}

Different from supervised VO learning \cite{turan2017endovo, turanendovmfuse2017, turanendosensor2017}, where camera poses and/or depth ground truths are required to train the neural network, the core idea underlying our unsupervised pose and depth prediction method is to make use of the view synthesis constraint as the supervision metric, which forces the neural network to synthesize target image from multiple source images acquired from different camera poses. This synthesis is performed using estimated depth image, estimated target camera pose values in 6-DoF and nearby color values from source images. In addition, a reliability mask is trained to detect sensor occlusions, non-rigid deformations of the soft organ tissue and lack of textures inside the explored organ. 

%%%%%%%%%%%%%%%%%%%%%%%%%%%%%%%%%%%%%%%%%%%%%%%%%%%%%%%%%%%%%%%%%%%%%%%%%%%%

\subsection{View synthesis as supervision metric}

To provide a supervision to the neural network, view synthesis is accomplished by training with consecutive images. As input, we take a sequence of 3 consecutive frames, and choose the middle frame as a target frame. Sequences are denoted by $<I_{t-1}, I_{t}, I_{t+1}>$ where $I_{t}$ is the target view and rest of images are source views $I_{s} = <I_{t-1}, I_{t+1}>$, which are used to render the target image (see Fig. \ref{fig:seq}). The objective function of the view synthesis is:
\begin{equation}
\mathcal{L}_{vs} = \sum_{s}\sum_p | I_t (p) - \hat{I}_s(p) |
\end{equation}
where $p$ is pixel coordinate, and $\hat{I}_s$ is the source view $I_s$ warped to the target view making use of the estimated depth image $\hat{D}_t$ and $4\times 4$ camera transformation matrix $\hat{T}_{t\rightarrow s}$  \cite{fehn2004depth}. Let $p_t$ represent the homogeneous pixel coordinates in the target view, and $K$ be the camera intrinsics matrix.

 $p_t$ is projected coordinate on the source view and $p_s$ is acquired by:
\begin{equation}
p_s \sim K \hat{T}_{t\rightarrow s} \hat{D}_t(p_t) K^{-1} p_t
\end{equation}

Note that the value of $p_s$ is not discrete. To find the expected intensity value at that position, bilinear interpolation among four discrete neighbors of $p_s$ is used \cite{zhou2016view}:
\begin{equation}
\hat{I}_s(p_t) = I_s(p_s) = \sum_{i\in\{top,bottom\}, j\in\{left,right\}}w^{ij}I_s(p_s^{ij})
\end{equation}

Let $w^{ij}$ be the proximity value between projected and neighboring pixels summing up to one and $\hat{I}_s$ be the estimated mean intensity for projected pixel $p_s$.

View synthesis approach assumes that camera sensor is not occluded, non-rigid deformations are avoided and explored organ surface obeys Lambertian surface rules enabling photometric error minimization between target and source views.
These assumptions are frequently violated in endoscopic type of videos:
\begin{enumerate}
	\item Sensor occlusions occur often due to peristaltic organ motions.
	\item Inner organs have in general a non-rigid structure meaning deformations cannot be completely avoided.
	\item Organ fluids cause specularities which violate the Lambertian surface rules.
\end{enumerate}
To overcome these, we trained a soft reliability mask which labels each target-source pixel pair as reliable to be used for view-synthesis or believed to violate assumptions because of being affected by occlusions, non-rigid deformations and/or specularities. Incorporating the soft-reliability mask  $\hat{E}_s$, the view synthesis equation is updated as:
\begin{equation}
    \mathcal{L}_{vs} = \sum_{<I_{t-1}, I_{t+1}> \in \mathcal{S}}\sum_p \hat{E}_s(p) | I_t (p) - \hat{I}_s(p) |~.
\label{eq:exp}
\end{equation}

Minimizing this energy function without regularizer will force mask to be zero across the whole image domain. To overcome this problem and obtain a reasonable mask, a regularization term is to use which describes the prior knowledge about reliability mask. Hence, let $\mathcal{L}_{reg}(\hat{E}_s^l)$ be the regularization term that minimizes the cross-entropy loss and prevents trivial solutions. 
Finally, since gradients are derived from differences between four neighbors and corresponding pixel intensities of source and target frames, a smoothness loss $\mathcal{L}^l_{smooth}$ is needed. The multiscale pyramid and smoothness loss for gradients are extracted from larger spatial regions. This leads to the following energy function:
\begin{equation}
\mathcal{L}_{final} = \sum_l \mathcal{L}_{vs}^l +  \lambda_s \mathcal{L}^l_{smooth} + \lambda_e \sum_{s} \mathcal{L}_{reg}(\hat{E}_s^l)~
\label{eq:final}
\end{equation}
Here, $s$ indexes source images, $l$ indexes images from different scales, $\lambda_s$ is the regularization weight for depth smoothness, and $\lambda_e$ is the weight for reliability mask. 

%%%%%%%%%%%%%%%%%%%%%%%%%%%%%%%%%%%%%%%%%%%%%%%%%%%%%%%%%%%%%%%%%%%%%%%%%%%%

\subsection{Network architecture}
\label{sec:network_arch}

\begin{figure*}[t]
    \centering
    \begin{subfigure}[t]{0.4\textwidth} 
    \includegraphics[width=\textwidth]{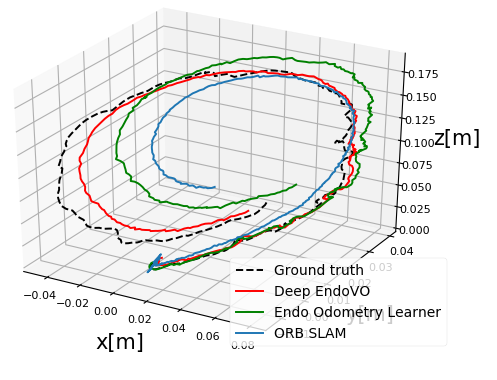}
    \caption{Trajectory 1}
    \label{fig:traj1} 
    \end{subfigure}
    ~
    \begin{subfigure}[t]{0.4\textwidth} 
    \includegraphics[width=\textwidth]{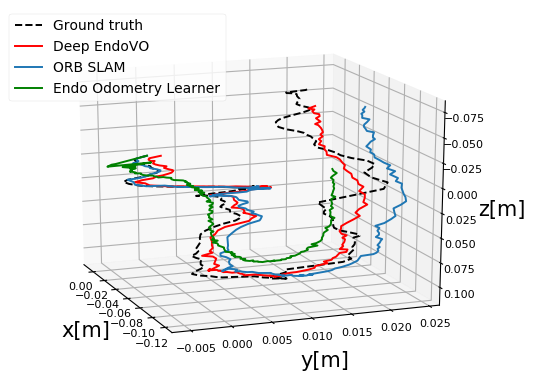}
    \caption{Trajectory 2}
    \label{fig:traj2} 
    \end{subfigure}
    ~
    \begin{subfigure}[t]{0.4\textwidth} 
    \includegraphics[width=\textwidth]{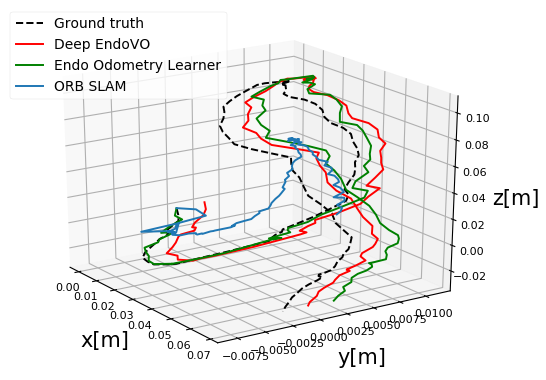}
    \caption{Trajectory 3}
    \label{fig:traj3}       % Give a unique label
    \end{subfigure}
    %\\
    ~
    \begin{subfigure}[t]{0.4\textwidth} 
    \includegraphics[width=\textwidth]{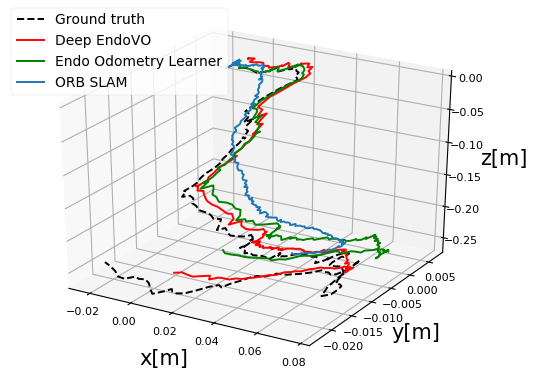}
    \caption{Trajectory 4}
    \label{fig:traj4} 
    \end{subfigure}
    ~
    
    \caption{Sample trajectories comparing the unsupervised learning method with ORB SLAM, EndoVO and OptiTrack ground truth in millimetric scale. Deep EndoVO shows the best odometry estimations, whereas ORB SLAM fails to track some fine-scale motions. Tracking performance of unsupervised odometry lies inbetween of ORB SLAM and Deep EndoVO; many fine-scale motions are successfully caught in detail, however there is still a certain amount of drift.  }
    \label{fig:trajs}
\end{figure*}

As mentioned earlier, our problem is estimating odometry in a textureless scene by using only sequenced RGB frames as input. Since classical methods fail to cope with this problem, we use DL methods where we get our motivation from recent works \cite{vijayanarasimhan17} and \cite{zhou2017unsupervised} which propose improvements by autoencoder based architectures. Our overall DL model as shown in Fig. \ref{fig:deeparchitecture} consists of two end-to-end frameworks.

The first architecture is employed to predict single-view depths by creating disparity map outputs. The encoder-decoder convolutional layers are followed by a prediction layer, whose outputs are constrained by $1/(\alpha * sigmoid(x)+\beta)$ with  $\alpha = 10$ and  $\beta = 0.1$ to ensure that predictions occur in a desirable interval. 

The second network tries to estimate relative pose, parameterized by SE(3) motions between views, and the reliability mask. The encoder part for pose estimation and reliability mask are same, where they share weights in the first five feature extractor convolutional layers and divide into two tracks afterwards. Pose is estimated by encoder's $6*(N-1)$ channels, as translation and rotation parameters. The decoder part consists of five deconvolutional layers and generates multiscale mask predictions. There are four output channels for each prediction layer, and each two of them predict the reliability for input source-target pairs by softmax normalization.

Both networks are trained and optimized jointly. On the other hand, both networks can be tested and evaluated independently. Testing and training pipelines are illustrated in Fig.  \ref{fig:networks}.

% For single-view depth prediction, we adopt the SfmLearner architecture proposed in \cite{zhou2017unsupervised} that is mainly based on an encoder-decoder design with skip connections and multi-scale side predictions. All convolution layers are followed by ReLU activation except for the prediction layers, where we use $1/(\alpha * sigmoid(x)+\beta)$ with  $\alpha = 10$ and  $\beta = 0.1$ to constrain the predicted depth to be always positive within a reasonable range. The input to the pose estimation network is the target view concatenated with all the source views along the color channels, and the outputs are the relative poses between the target view and each of the source views. The network consists of $7$ stride-$2$ convolutions followed by a $1×1$ convolution with $6*(N-1)$ output channels (corresponding to $3$ Euler angles and 3-D translation for each source view). Finally, global average pooling is applied to aggregate predictions at all spatial locations. The reliability mask prediction network shares the first five feature encoding layers with the pose network, followed by five deconvolution layers with multi-scale side predictions. The number of output channels for each prediction layer is $2*(N-1)$, with every two channels normalized by softmax to obtain the reliability prediction for the corresponding source-target pair.

%%%%%%%%%%%%%%%%%%%%%%%%%%%%%%%%%%%%%%%%%%%%%%%%%%%%%%%%%%%%%%%%%%%%%%%%%%%%%%%%%%%%%%%%%%%%%%%%%%%%%%%%%%%%%%%%%%%%%%%%%%%%%%%%%%%%%%%%%%%%%%%%%%%%%%%%%%%%%%%%

\section{EVALUATION AND RESULTS}
\label{sec:results}

\subsection{Dataset and Transfer learning}
We used transfer learning to have an initialization for neural network weights since we lack huge amounts of labeled data. For pretraining, DL model proposed by Zhou et al.\cite{zhou2017unsupervised} is employed. The model is implemented with publicly available Tensorflow framework and pretrained with the KITTI dataset. Batch normalization is used for all of the layers except the outputs. Adam optimization is chosen to increase the convergence rate, with $\beta_1 = 0.9$, $\beta_2 = 0.999$, learning rate of $0.1$ and mini-batch size of $8$. We used the model which was trained with $50K$ images and converged after 150K iterations. The model requires sequential images with size 128 x 416. On top of the model pretrained by a KITTI dataset, we fine-tuned the architecture with our domain data from endoscopic capsule robot by employing a GeForce GTX 1070 model GPU. Our dataset was collected in an experimental setup for an ex-vivo parcine stomach shown in Fig. \ref{fig:opti} and it contains $12K$ frames with ground truth odometry obtained by OptiTrack visual tracking system. In this experiment, we fix the length of input image sequences into three frames. We used $10K$ frames for training, $1K$ for cross validation and $1K$ for evaluation and testing.

%We implemented the system using the publicly available TensorFlow framework. For all the experiments, we set $\lambda_s = 0.4/l$ ($l$ is the downscaling factor for the corresponding scale) and $\lambda_e = 0.15$. During training, we used batch normalization for all the layers except for the output layers, and the Adam optimizer with $\beta_1 = 0.9$, $\beta_2 = 0.999$, learning rate of $0.0003$ and mini-batch size of $8$. The training typically converges after about $120$K iterations. Experiments are performed with image sequences captured with four different endoscopic monocular camera. We resize the images to $300 × 300$ during training. We train our system on our pig stomach datasets. To increase the generalizability of the model and to reduce the training time, the weights of the network are initialized with the model that is trained on KITTI dataset for visual odometry task. Our dataset consists of $50$K sequences, out of which we use $40$K for training and $10$K for validation. To evaluate the performance of our pose estimation network, we applied our system to the pig stomach dataset that contains $10$K frames with ground truth odometry obtained by OptiTrack visual tracking system. In this experiment, we fix the length of input image sequences to our system to 7 frames. 

%%%%%%%%%%%%%%%%%%%%%%%%%%%%%%%%%%%%%%%%%%%%%%%%%%%%%%%%%%%%%%%%%%%%%%%%%%%%

\subsection{Pose estimation and Odometry benchmark}
Our pose estimation network is tested with $1K$ frames. The network outputs the pose predictions as 6-DoF motion (Euclidean coordinates for translation and rotation) between sequences. Ground truth data was established with the OptiTrack mechanism. %We have used TUM Benchmark's odometry evaluation to compare our results with ground truth\ref{sturm12iros}. 
Some examples from odometry outputs can be seen in Fig. \ref{fig:trajs}. Here, we illustrate only short sequences qualitatively. It can be seen that the main trajectory results successfully differentiate the major displacements with a minor amount of drift.

\begin{figure*}[t]
\centering
\begin{subfigure}[t]{0.48\textwidth} 
\includegraphics[width=\textwidth, height=1.5in]{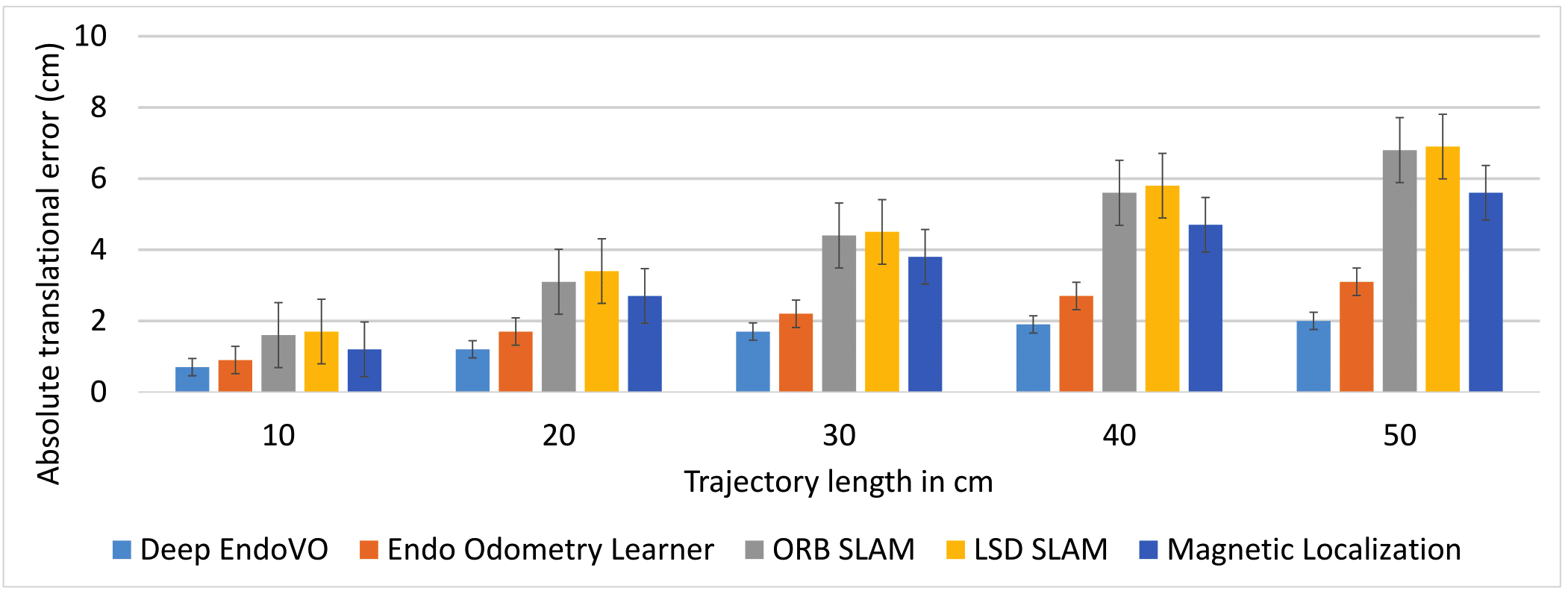}
        \caption{Translational error results}
\label{fig:err_trans}
\end{subfigure}
~
\begin{subfigure}[t]{0.48\textwidth} 
\includegraphics[width=\textwidth, height=1.5in]{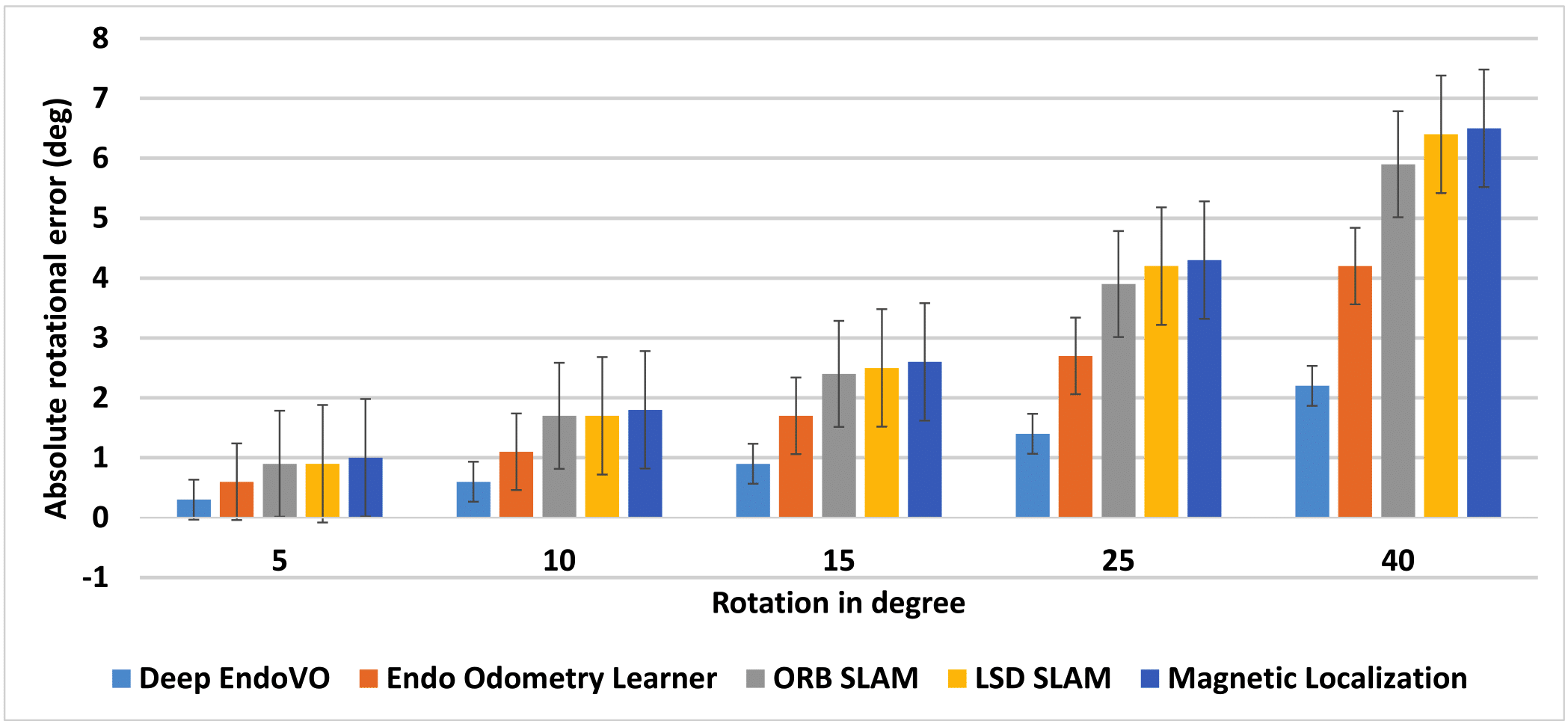}
        \caption{Rotational error results}
        \label{fig:err_rot}
\end{subfigure}
~

\caption{Translational (a) and rotational (b) error results for ORB SLAM, LSD SLAM, Deep EndoVO, magnetic localization and our proposed supervised method. It is clear that in both rotational and translational motions, our unsupervised odometry outperforms ORB SLAM, LSD SLAM and magnetic localization, whereas Deep EndoVO shows best performance. For example, for trajectory length of 10 cm, Deep EndoVO and our method results in a translational error less than 1 cm, and others are slightly above 1 cm. In terms of rotational motion, a 5 degree change has an effect of less than 1 degree in Deep EndoVO and our method, however rest of the methods are closer to 1 degree. Translational results indicate that the proposed method shows robustness for increasing trajectory lengths and remains close to the ground truth trajectory. The trajectory length increase from 10 cm to 50 cm results a change of more than 4 cm in both ORB SLAM and LSD SLAM methods, whereas our error increases around 1 cm. }
\label{fig:err}
\end{figure*}

\begin{figure}
\includegraphics[width=\columnwidth]{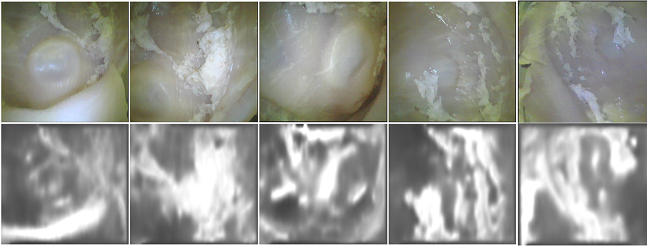}
\caption{Sample disparity map estimations from ex-vivo porcine stomach dataset. Even though depth estimations lack fine-scale details in low textured areas, major depth differences were successfully caught. }
\label{fig:res_depth}
\end{figure} 

We compare our ego-motion estimation method with monocular ORB-SLAM \cite{mur2015orb}, Deep EndoVO \cite{turan2017endovo}, LSD SLAM \cite{engel2014lsd} using Absolute Trajectory Error (ATE) \cite{mur2015orb} for the alignment with the ground truth.
As shown in Fig. \ref{fig:trajs} and error bars in Fig. \ref{fig:err_trans}, \ref{fig:err_rot}, our method outperforms ORB SLAM and LSD SLAM which are state-of-the art widely used SLAM methods. Because of the geometric and photometric properties of scenes, these methods fail to find and match proper keypoints. Magnetic localization also outperforms ORB-SLAM and LSD-SLAM, because magnetic localization does not depend on textural geometry of the scene. Even though the proposed method is unsupervised, its translational and rotational accuracies are comparable with Deep EndoVO approach which is a supervised odometry learning method.
 
\subsection{Depth Estimation}
\begin{figure}[t]

	\begin{subfigure}[t]{0.15\textwidth} 
	    \includegraphics[width=1\linewidth]{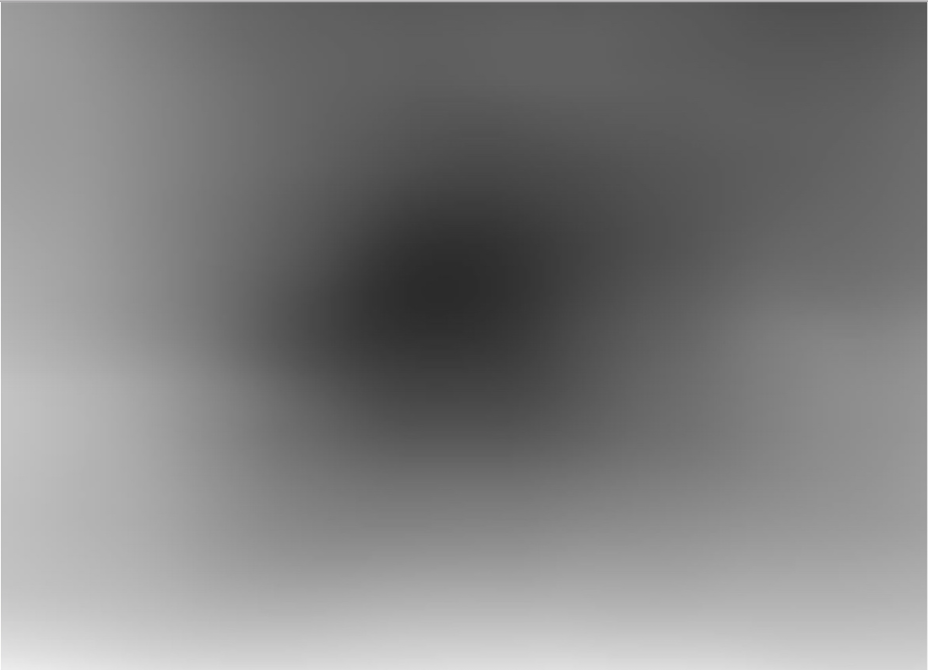}
        \caption{Without fine-tuning (KITTI)} \label{fig:a}
	\end{subfigure}
	~
	\begin{subfigure}[t]{0.15\textwidth} 
	    \includegraphics[width=1\linewidth]{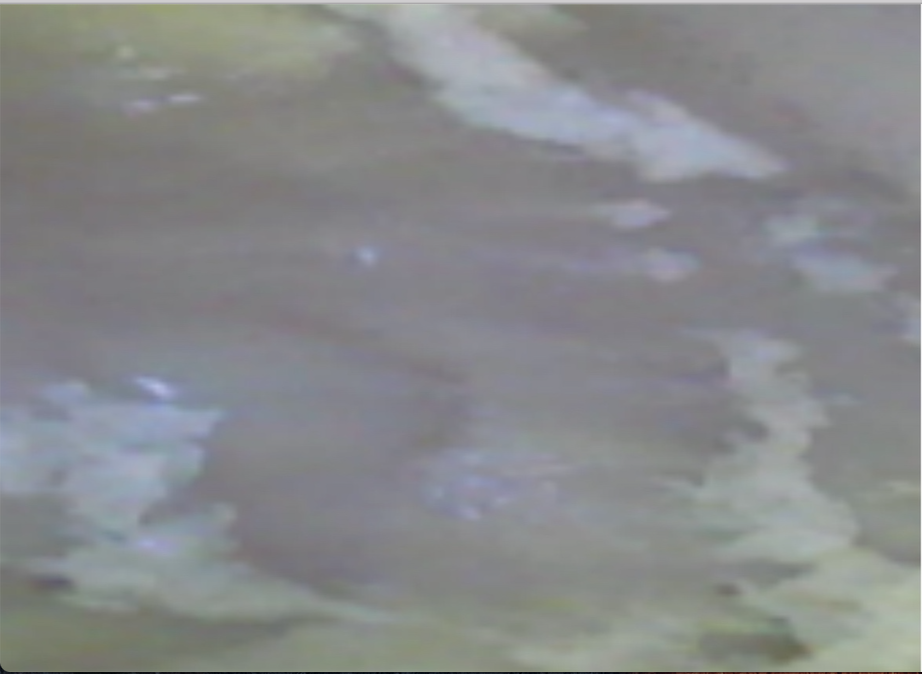}
        \caption{Original image} \label{fig:b}
	\end{subfigure}
	~
	\begin{subfigure}[t]{0.15\textwidth} 
	    \includegraphics[width=1\linewidth]{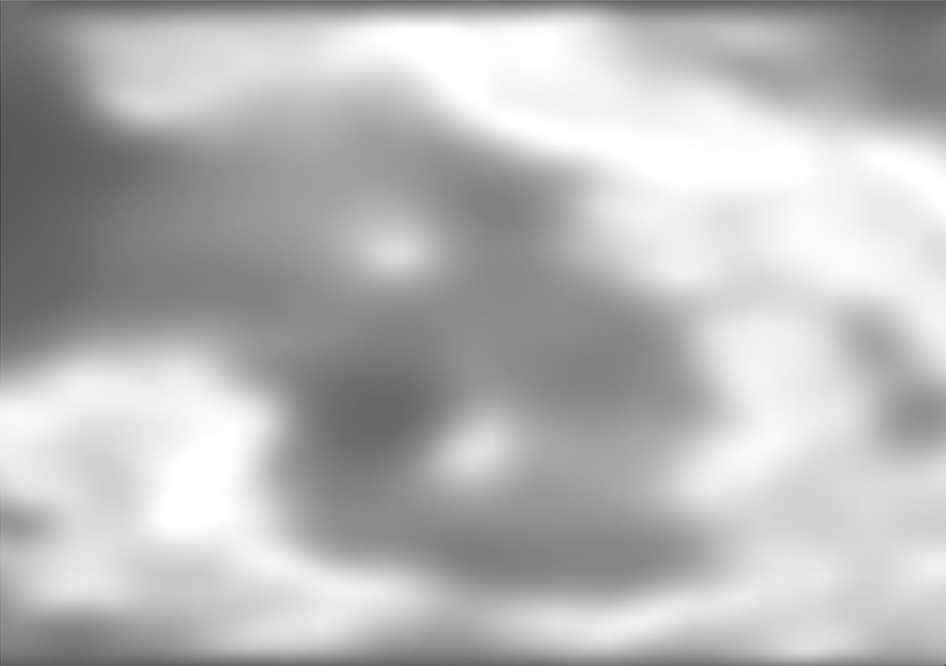}
        \caption{After transfer learning} \label{fig:c}
	\end{subfigure}
	
	\caption{Disparity map outputs before and after fine-tuning on top of KITTI. (a) shows the estimation without fine-tuning. Since there is no object in front of the camera in KITTI images, the resulting disparity maps have a dark region in the center. Moreover, the disparity map has a poor quality. After transfer learning and training with porcine stomach dataset in addition to KITTI images, the quality of the disparity map drastically increases and the dark hole in the center of the image dissapears (c). }
    \label{fig:finetune}
\end{figure}

The neural network model creates depth estimation as a disparity map for a given view. Some estimation results can be seen in Fig. \ref{fig:res_depth}. It is clear that major depth differences are captured by the network. However, since stomach surface is non-Lambertian and the light source is attached to camera, it becomes more challenging to reproduce a robust algorithm. In the disparity map output of the network, it is observable that there are minor errors at some low textured regions or on high gradient parts such as sharp edges. However, our improvement on overall depth estimation with fine-tuning can be seen in Fig. \ref{fig:finetune}.

%%%%%%%%%%%%%%%%%%%%%%%%%%%%%%%%%%%%%%%%%%%%%%%%%%%%%%%%%%%%%%%%%%%%%%%%%%%%%%%%%%%%%%%%%%%%%%%%%%%%%%%%%%%%%%%%%%%%%%%%%%%%%%%%%%%%%%%%%%%%%%%%%%%%%%%%%%%%%%%%

\section{CONCLUSIONS}
\label{sec:conclusion}

In this paper we applied unsupervised DL method for estimating VO and depth for endoscopic capsule robot videos. Even though our method performs comparably well to supervised EndoVO method and outperforms existing state of the arts SLAM algorithms ORB and LSD SLAM, some playroom for the improvements of the method still remains: 
\begin{itemize}
    \item Accuracy of the results can be improved by increasing sequence size of inputs. As well, additional training data generated by augmentation techniques could improve the performance of the method for cases where non-rigid deformations, occlusions and heavy specularities exist.
    \item Since our capsule robot also uses rolling shutter camera, instead using KITTI dataset captured by global shutter camera, we could also incorporate Cityscapes dataset captured by rolling shutter camera.
    \item The quality of estimated depth maps can be improved by combining the depth output of our method with shading based depth estimation. In that way, a  more realistic and therapeutically relevant 3D reconstruction of the explored inner organ could be achieved.
    \item The dependency of the proposed method on the camera intrinsics matrix makes it rather impractical to be used for random videos streaming from hospitals with unknown calibration matrix. 
    \item It would be interesting to extend our network to perform further tasks such as tissue segmentation and disease detection. 
\end{itemize}

%                                  % on the last page of the document manually. It shortens
%                                  % the textheight of the last page by a suitable amount.
%                                  % This command does not take effect until the next page
%                                  % so it should come on the page before the last. Make
%                                  % sure that you do not shorten the textheight too much.

%%%%%%%%%%%%%%%%%%%%%%%%%%%%%%%%%%%%%%%%%%%%%%%%%%%%%%%%%%%%%%%%%%%%%%%%%%%%%%%%

%%%%%%%%%%%%%%%%%%%%%%%%%%%%%%%%%%%%%%%%%%%%%%%%%%%%%%%%%%%%%%%%%%%%%%%%%%%%%%%

%%%%%%%%%%%%%%%%%%%%%%%%%%%%%%%%%%%%%%%%%%%%%%%%%%%%%%%%%%%%%%%%%%%%%%%%%%%%%%%
%\section*{APPENDIX}

%\section*{ACKNOWLEDGMENT}

%%%%%%%%%%%%%%%%%%%%%%%%%%%%%%%%%%%%%%%%%%%%%%%%%%%%%%%%%%%%%%%%%%%%%%%%%%%%%%%%

\bibliographystyle{IEEEtran}
\bibliography{mybibfile}

\begin{thebibliography}{10}
\providecommand{\url}[1]{#1}
\csname url@rmstyle\endcsname
\providecommand{\newblock}{\relax}
\providecommand{\bibinfo}[2]{#2}
\providecommand\BIBentrySTDinterwordspacing{\spaceskip=0pt\relax}
\providecommand\BIBentryALTinterwordstretchfactor{4}
\providecommand\BIBentryALTinterwordspacing{\spaceskip=\fontdimen2\font plus
\BIBentryALTinterwordstretchfactor\fontdimen3\font minus
  \fontdimen4\font\relax}
\providecommand\BIBforeignlanguage[2]{{%
\expandafter\ifx\csname l@#1\endcsname\relax
\typeout{** WARNING: IEEEtran.bst: No hyphenation pattern has been}%
\typeout{** loaded for the language `#1'. Using the pattern for}%
\typeout{** the default language instead.}%
\else
\language=\csname l@#1\endcsname
\fi
#2}}

\bibitem{sitti2015biomedical}
M.~Sitti, H.~Ceylan, W.~Hu, J.~Giltinan, M.~Turan, S.~Yim, and E.~Diller,
  ``Biomedical applications of untethered mobile milli/microrobots,''
  \emph{Proceedings of the IEEE}, vol. 103, no.~2, pp. 205--224, 2015.

\bibitem{turan2017endovo}
M.~Turan, Y.~Almalioglu, H.~Araujo, E.~Konukoglu, and M.~Sitti, ``Deep endovo:
  A recurrent convolutional neural network (rcnn) based visual odometry
  approach for endoscopic capsule robots,'' \emph{arXiv preprint
  arXiv:1708.06822}, 2017.

\bibitem{goenka2014capsule}
M.~K. Goenka, S.~Majumder, and U.~Goenka, ``Capsule endoscopy: Present status
  and future expectation,'' \emph{World J Gastroenterol}, vol.~20, no.~29, pp.
  10\,024--10\,037, 2014.

\bibitem{turanendovmfuse2017}
\BIBentryALTinterwordspacing
M.~Turan, Y.~Almalioglu, H.~Gilbert, A.~E. Sari, U.~Soylu, and M.~Sitti,
  ``Endo-vmfusenet: Deep visual-magnetic sensor fusion approach for
  uncalibrated, unsynchronized and asymmetric endoscopic capsule robot
  localization data,'' \emph{CoRR}, vol. abs/1709.06041, 2017. [Online].
  Available: \url{http://arxiv.org/abs/1709.06041}
\BIBentrySTDinterwordspacing

\bibitem{TURAN20181861}
\BIBentryALTinterwordspacing
M.~Turan, Y.~Almalioglu, H.~Araujo, E.~Konukoglu, and M.~Sitti, ``Deep endovo:
  A recurrent convolutional neural network (rcnn) based visual odometry
  approach for endoscopic capsule robots,'' \emph{Neurocomputing}, vol. 275,
  pp. 1861 -- 1870, 2018. [Online]. Available:
  \url{http://www.sciencedirect.com/science/article/pii/S092523121731665X}
\BIBentrySTDinterwordspacing

\bibitem{turanendosensor2017}
\BIBentryALTinterwordspacing
M.~Turan, Y.~Almalioglu, H.~Gilbert, H.~Ara{\'{u}}jo, T.~Cemgil, and M.~Sitti,
  ``Endosensorfusion: Particle filtering-based multi-sensory data fusion with
  switching state-space model for endoscopic capsule robots,'' \emph{CoRR},
  vol. abs/1709.03401, 2017. [Online]. Available:
  \url{http://arxiv.org/abs/1709.03401}
\BIBentrySTDinterwordspacing

\bibitem{Turan2017}
\BIBentryALTinterwordspacing
M.~Turan, Y.~Almalioglu, H.~Araujo, E.~Konukoglu, and M.~Sitti, ``A non-rigid
  map fusion-based direct slam method for endoscopic capsule robots,''
  \emph{International Journal of Intelligent Robotics and Applications},
  vol.~1, no.~4, pp. 399--409, Dec 2017. [Online]. Available:
  \url{https://doi.org/10.1007/s41315-017-0036-4}
\BIBentrySTDinterwordspacing

\bibitem{Turan2018}
\BIBentryALTinterwordspacing
M.~Turan, Y.~Y. Pilavci, I.~Ganiyusufoglu, H.~Araujo, E.~Konukoglu, and
  M.~Sitti, ``Sparse-then-dense alignment-based 3d map reconstruction method
  for endoscopic capsule robots,'' \emph{Machine Vision and Applications},
  vol.~29, no.~2, pp. 345--359, Feb 2018. [Online]. Available:
  \url{https://doi.org/10.1007/s00138-017-0905-8}
\BIBentrySTDinterwordspacing

\bibitem{nakamura2008capsule}
T.~Nakamura and A.~Terano, ``Capsule endoscopy: past, present, and future,''
  \emph{Journal of gastroenterology}, vol.~43, no.~2, pp. 93--99, 2008.

\bibitem{munoz2014review}
F.~Munoz, G.~Alici, and W.~Li, ``A review of drug delivery systems for capsule
  endoscopy,'' \emph{Advanced drug delivery reviews}, vol.~71, pp. 77--85,
  2014.

\bibitem{carpi2011magnetically}
F.~Carpi, N.~Kastelein, M.~Talcott, and C.~Pappone, ``Magnetically controllable
  gastrointestinal steering of video capsules,'' \emph{IEEE Transactions on
  Biomedical Engineering}, vol.~58, no.~2, pp. 231--234, 2011.

\bibitem{keller2012method}
H.~Keller, A.~Juloski, H.~Kawano, M.~Bechtold, A.~Kimura, H.~Takizawa, and
  R.~Kuth, ``Method for navigation and control of a magnetically guided capsule
  endoscope in the human stomach,'' in \emph{Biomedical Robotics and
  Biomechatronics (BioRob), 2012 4th IEEE RAS \& EMBS International Conference
  on}.\hskip 1em plus 0.5em minus 0.4em\relax IEEE, 2012, pp. 859--865.

\bibitem{mahoney2013managing}
A.~W. Mahoney, S.~E. Wright, and J.~J. Abbott, ``Managing the attractive
  magnetic force between an untethered magnetically actuated tool and a
  rotating permanent magnet,'' in \emph{Robotics and Automation (ICRA), 2013
  IEEE International Conference on}.\hskip 1em plus 0.5em minus 0.4em\relax
  IEEE, 2013, pp. 5366--5371.

\bibitem{yim2014biopsy}
S.~Yim, E.~Gultepe, D.~H. Gracias, and M.~Sitti, ``Biopsy using a magnetic
  capsule endoscope carrying, releasing, and retrieving untethered
  microgrippers,'' \emph{IEEE Transactions on Biomedical Engineering}, vol.~61,
  no.~2, pp. 513--521, 2014.

\bibitem{petruska2013omnidirectional}
A.~J. Petruska and J.~J. Abbott, ``An omnidirectional electromagnet for remote
  manipulation,'' in \emph{Robotics and Automation (ICRA), 2013 IEEE
  International Conference on}.\hskip 1em plus 0.5em minus 0.4em\relax IEEE,
  2013, pp. 822--827.

\bibitem{DBLP:journals/corr/TuranAKS17}
\BIBentryALTinterwordspacing
M.~Turan, Y.~Almalioglu, E.~Konukoglu, and M.~Sitti, ``A deep learning based 6
  degree-of-freedom localization method for endoscopic capsule robots,''
  \emph{CoRR}, vol. abs/1705.05435, 2017. [Online]. Available:
  \url{http://arxiv.org/abs/1705.05435}
\BIBentrySTDinterwordspacing

\bibitem{DBLP:journals/corr/TuranPJAKS17}
\BIBentryALTinterwordspacing
M.~Turan, Y.~Y. Pilavci, R.~Jamiruddin, H.~Ara{\'{u}}jo, E.~Konukoglu, and
  M.~Sitti, ``A fully dense and globally consistent 3d map reconstruction
  approach for {GI} tract to enhance therapeutic relevance of the endoscopic
  capsule robot,'' \emph{CoRR}, vol. abs/1705.06524, 2017. [Online]. Available:
  \url{http://arxiv.org/abs/1705.06524}
\BIBentrySTDinterwordspacing

\bibitem{DBLP:journals/corr/TuranAJAKS17}
\BIBentryALTinterwordspacing
M.~Turan, A.~Abdullah, R.~Jamiruddin, H.~Ara{\'{u}}jo, E.~Konukoglu, and
  M.~Sitti, ``Six degree-of-freedom localization of endoscopic capsule robots
  using recurrent neural networks embedded into a convolutional neural
  network,'' \emph{CoRR}, vol. abs/1705.06196, 2017. [Online]. Available:
  \url{http://arxiv.org/abs/1705.06196}
\BIBentrySTDinterwordspacing

\bibitem{DBLP:journals/corr/TuranAAKS17}
\BIBentryALTinterwordspacing
M.~Turan, Y.~Almalioglu, H.~Ara{\'{u}}jo, E.~Konukoglu, and M.~Sitti, ``A
  non-rigid map fusion-based rgb-depth {SLAM} method for endoscopic capsule
  robots,'' \emph{CoRR}, vol. abs/1705.05444, 2017. [Online]. Available:
  \url{http://arxiv.org/abs/1705.05444}
\BIBentrySTDinterwordspacing

\bibitem{szeliski1999prediction}
R.~Szeliski, ``Prediction error as a quality metric for motion and stereo,'' in
  \emph{Computer Vision, 1999. The Proceedings of the Seventh IEEE
  International Conference on}, vol.~2.\hskip 1em plus 0.5em minus 0.4em\relax
  IEEE, 1999, pp. 781--788.

\bibitem{zhou2017unsupervised}
T.~Zhou, M.~Brown, N.~Snavely, and D.~G. Lowe, ``Unsupervised learning of depth
  and ego-motion from video,'' in \emph{CVPR}, vol.~2, no.~6, 2017, p.~7.

\bibitem{flynn2016deepstereo}
J.~Flynn, I.~Neulander, J.~Philbin, and N.~Snavely, ``Deepstereo: Learning to
  predict new views from the world's imagery,'' in \emph{Proceedings of the
  IEEE Conference on Computer Vision and Pattern Recognition}, 2016, pp.
  5515--5524.

\bibitem{than2012review}
T.~D. Than, G.~Alici, H.~Zhou, and W.~Li, ``A review of localization systems
  for robotic endoscopic capsules,'' \emph{IEEE Transactions on Biomedical
  Engineering}, vol.~59, no.~9, pp. 2387--2399, 2012.

\bibitem{fluckiger2007ultrasound}
M.~Fluckiger and B.~J. Nelson, ``Ultrasound emitter localization in
  heterogeneous media,'' in \emph{2007 29th Annual International Conference of
  the IEEE Engineering in Medicine and Biology Society}.\hskip 1em plus 0.5em
  minus 0.4em\relax IEEE, 2007, pp. 2867--2870.

\bibitem{rubin2006sonographic}
J.~M. Rubin, H.~Xie, K.~Kim, W.~F. Weitzel, S.~Y. Emelianov, S.~R. Aglyamov,
  T.~W. Wakefield, A.~G. Urquhart, and M.~OâDonnell, ``Sonographic
  elasticity imaging of acute and chronic deep venous thrombosis in humans,''
  \emph{Journal of Ultrasound in Medicine}, vol.~25, no.~9, pp. 1179--1186,
  2006.

\bibitem{kim2008noninvasive}
K.~Kim, L.~A. Johnson, C.~Jia, J.~C. Joyce, S.~Rangwalla, P.~D. Higgins, and
  J.~M. Rubin, ``Noninvasive ultrasound elasticity imaging (uei) of crohn's
  disease: animal model,'' \emph{Ultrasound in medicine \& biology}, vol.~34,
  no.~6, pp. 902--912, 2008.

\bibitem{yim20133}
S.~Yim and M.~Sitti, ``3-d localization method for a magnetically actuated soft
  capsule endoscope and its applications,'' \emph{IEEE Transactions on
  Robotics}, vol.~29, no.~5, pp. 1139--1151, 2013.

\bibitem{mayerIHFCDB15}
\BIBentryALTinterwordspacing
N.~Mayer, E.~Ilg, P.~H{\"{a}}usser, P.~Fischer, D.~Cremers, A.~Dosovitskiy, and
  T.~Brox, ``A large dataset to train convolutional networks for disparity,
  optical flow, and scene flow estimation,'' \emph{CoRR}, vol. abs/1512.02134,
  2015. [Online]. Available: \url{http://arxiv.org/abs/1512.02134}
\BIBentrySTDinterwordspacing

\bibitem{fehn2004depth}
C.~Fehn, ``Depth-image-based rendering (dibr), compression, and transmission
  for a new approach on 3d-tv,'' in \emph{Stereoscopic Displays and Virtual
  Reality Systems XI}, vol. 5291.\hskip 1em plus 0.5em minus 0.4em\relax
  International Society for Optics and Photonics, 2004, pp. 93--105.

\bibitem{zhou2016view}
T.~Zhou, S.~Tulsiani, W.~Sun, J.~Malik, and A.~A. Efros, ``View synthesis by
  appearance flow,'' \emph{CoRR}, vol. abs/1605.03557, 2016.

\bibitem{vijayanarasimhan17}
S.~Vijayanarasimhan, S.~Ricco, C.~Schmid, R.~Sukthankar, and K.~Fragkiadaki,
  ``Sfm-net: Learning of structure and motion from video,'' \emph{CoRR}, vol.
  abs/1704.07804, 2017.

\bibitem{mur2015orb}
R.~Mur-Artal, J.~Montiel, and J.~D. Tard{\'o}s, ``Orb-slam: a versatile and
  accurate monocular slam system,'' \emph{IEEE Transactions on Robotics},
  vol.~31, no.~5, pp. 1147--1163, 2015.

\bibitem{engel2014lsd}
J.~Engel, T.~Sch{\"o}ps, and D.~Cremers, ``Lsd-slam: Large-scale direct
  monocular slam,'' in \emph{European Conference on Computer Vision}.\hskip 1em
  plus 0.5em minus 0.4em\relax Springer, 2014, pp. 834--849.

\end{thebibliography}
\end{document}